\documentclass[runningheads]{llncs}
\usepackage[T1]{fontenc}
\usepackage{graphicx}
\usepackage{caption}
\usepackage{subcaption}
\usepackage{booktabs}
\usepackage{tabularx}
\usepackage{makecell}
\begin{document}
%
\title{Evolving Features vs Evolving Entire Trees with GP for Interpretable Survival Analysis}
\titlerunning{GP for Interpretable Survival Analysis}
%
\author{Thalea Schlender \inst{1}\orcidID{0000-0001-7385-112X} \and
Peter A.N. Bosman\inst{2}\orcidID{0000-0002-4186-6666} \and
Tanja Alderliesten\inst{1}\orcidID{0000-0003-4261-7511}}
\authorrunning{T.Schlender et al.}
%
\institute{Leiden University Medical Center, Leiden, The Netherlands\and
Centrum Wiskunde \& Informatica, Amsterdam, The Netherlands.}
\maketitle              
\begin{abstract}
Survival analysis concerns the task of predicting the time until an event occurs. Often used in the medical field, survival analysis deals with incomplete (i.e., censored) data, for instance, from patients who did not experience the event during the duration of the study. For practical use, both accuracy and interpretability are important.

Survival trees are easy-to-follow survival models that split the patient cohort recursively into discrete patient groups. Whilst survival trees can capture complex relationships in survival problems, to do so accurately, they typically need to grow large, threatening interpretability. Moreover, survival trees are often built using greedy approaches that may overlook globally optimal split combinations, limiting predictive performance.

Shallow survival trees require expressive, higher‑order feature combinations to achieve competitive accuracy. We therefore use genetic programming to multi‑objectively evolve inherently inspectable feature sets and study how they interact with different tree induction strategies. We further introduce an evolutionary approach that jointly optimises the survival tree structure and the non‑linear split logic.

Our findings demonstrate that evolutionary feature construction improves predictive performance across different tree induction strategies on two real-world datasets and two different survival tree depths. Given its speed and flexible presentation, the multi-objective evolution of entire trees holds the most future promise. 

\end{abstract}

\section{Introduction}
Predicting when a patient may experience a clinical event, such as relapse or death, is essential for guiding medical decisions and planning care. Survival analysis concerns the field of methods that predict survival and are specifically designed to handle censored data. Censored data consists of patients for whom only partial information is available since they, for instance, did not experience the event or dropped out during the duration of a medical study. 

One of the most widely used methods in survival analysis is Cox regression~\cite{cox1972regression}, typically defined through the proportional hazards model. Here, the hazard function is expressed as the product of a baseline hazard, which depends only on time, and an exponential function of a linear combination of covariates and their corresponding coefficients. Crucially, the covariate effects are assumed to be time-independent. As a result, Cox regression is limited in its ability to capture complex, time-varying, or non-linear relationships in the data. Further, the model relies on the proportional hazards assumption, which states that hazard ratios between individuals remain constant over time, i.e. if one patient has a higher hazard than another at one time point, this ordering and relative difference remain the same at all times. This assumption is often not met in practice.  

To model more complex covariate interactions, deep learning approaches have been proposed, a prominent example of which is deepSurv~\cite{katzman2018deepsurv}. Whilst deep learning based survival analysis models achieve state-of-the-art performance, these methods tend to be too complex to understand, limiting their practical use. Moreover, DeepSurv is essentially a neural‑network parametrisation of the Cox proportional hazards model, binding it to the proportional hazards assumption.

Interpretability is of importance for the practical adoption of survival models. Clinicians must be able to verify model behaviour, trace predictions back to patient characteristics, and extract clinically meaningful insights about risk factors. Such transparency not only supports trust in model predictions but also facilitates knowledge discovery and can advance clinical research.

To address interpretability in deep learning survival analysis models, \cite{knottenbelt2025coxkan} introduces CoxKAN,  which optimises the Cox loss within Kolmogorov-Arnold Networks~\cite{liu2024kan}.  Each node in the network is then represented by B-splines. Once trained, CoxKAN approximates these B-Splines via symbolic regression. Whilst powerful, the resulting expressions approximate the learned network, rather than being optimised for the survival prediction directly.  Further, CoxKAN is still bound by the Cox proportional hazards assumption. 

An intuitive and inherently interpretable survival analysis technique that does not rely on the proportional hazards assumption is survival trees~\cite{leblanc1993survivaltree}. These models recursively partition the patient cohort based on covariates by most commonly using univariate splitting rules, i.e. $x<t$, where $x$ is a covariate and $t$ is a given threshold. Survival trees thus aim to create subgroups with distinct survival characteristics by minimising an impurity or heterogeneity criterion. Each leaf in the survival tree is represented by a survival function from a non-parametric estimator calculated over the leaf's patient subgroup, which can then be used to predict survival for new patients. 

Survival trees can naturally capture non-linear relationships and interactions between covariates, but typically need to grow large to capture complex relationships between patient characteristics and survival. Whilst shallow survival trees are easy to follow and comprehend, survival trees of larger depth become difficult to comprehend and increase the risk of overfitting. 

Classical survival tree induction follows a greedy, top-down procedure~\cite{leblanc1993survivaltree}. At each node, the covariate and threshold that maximally reduce impurity for the remaining subset of patients are determined. While efficient, the tree structure is not optimised globally. To address this limitation, several non-greedy optimisation strategies have been proposed. One such approach is by \cite{kretowska2024global}, who investigated evolving oblique survival trees by jointly evolving both the survival tree structure and a linear combination of covariates in each node. Two major limitations are the use of linear combinations only and the fact that the performance of the survival trees is optimised single-objectively by adding a regularisation term to weigh the number of leaves in the survival tree. Other approaches result in provably optimal survival trees by leveraging dynamic programming~\cite{huisman2024optimal,zhang2024optimal}. However, these methods can only handle a limited set of discrete split conditions.

Despite such advances in survival tree induction, modelling complex covariate relationships may still require large trees because node splits are often limited to a single covariate or simple interactions. To mitigate this, non-linear feature construction methods can be used to enable shallow survival trees to be accurate. Limited work has been done to this end. In recent work, we introduced~\emph{PISA}, a Pipeline for Interpretable Survival Analysis, which enhances shallow survival trees through multiple-feature multi-objective feature construction~\cite{schlender2025pisa}. To effectively search for symbolic expressions as features, a multi-objective variant of the state-of-the-art genetic programming (GP) method GP-GOMEA~\cite{virgolin2020explaining,sijben2022multi} is used. PISA was shown to improve the predictive performance of shallow survival trees while maintaining interpretability. However, PISA has so far been limited to using constructed features together with greedy tree induction only.

The benefit of using a multi-objective approach to feature construction is that no specific balance between predictive performance and model complexity needs to be made a priori. Instead, a set of solutions is yielded, allowing users such as medical researchers or clinicians to select a model that best aligns with their needs, domain expertise, and clinical intuition. By creating inherently interpretable features for shallow survival trees, each presented model can moreover be individually inspected and validated.

  Building on the multi-objective feature construction step in PISA, this work makes three key contributions. First, we systematically investigate evolving non-linear feature sets for three different tree induction strategies, namely, a greedy tree induction originally used in PISA, an optimal tree induction strategy from \cite{huisman2024optimal}, and our new evolutionary approach. Second, we propose a novel evolutionary survival tree (Evolutionary ST) that jointly optimises tree structure, feature selection, and split points directly from the genotype, enabling multi-objective optimisation of the entire survival tree rather than only the constructed features. Finally, we enhance the evolutionary ST by proposing an initialisation strategy and adding GP-tree swapping to promote diversity and evolutionary search. 
  
We first analyse performance on a synthetic XOR problem designed to assess the ability to capture complex covariate interactions and subsequently consider two clinical benchmark datasets. We illustrate that multi-objective feature construction is key to making shallow survival trees powerful, while the multi-objective evolution of entire survival trees holds the most promise. 

\section{PISA}
In this work, we investigate multiple-feature multi-objective feature construction for shallow survival trees using different tree induction strategies. To embed our work as an extension to PISA, we reiterate PISA's methodology, but refer the reader to \cite{schlender2025pisa} for an in-depth description. PISA is a model-agnostic pipeline, which requires an elementary survival analysis approach (i.e. survival trees in this work) to be specified. 
\paragraph{Multiple feature multi-objective feature construction}
PISA uses GP to engineer feature sets for the elementary survival analysis approach. Specifically, multi-tree multi-objective GP-GOMEA~\cite{virgolin2020explaining,sijben2022multi,schlender2024improving} is used, as GP-GOMEA is highly adept at finding small, but accurate expressions~\cite{virgolin2020explaining}. The fitness of a feature set is the predictive performance of the elementary survival analysis model using said feature set, as well as the size of all features within the set included in the model. To acquire a robust estimate of the model's performance, the predictive ability is measured over 25 shuffle splits. Due to the inherent variability in evolutionary algorithms, as well as the inherent instability of survival analysis problems (with limited data), the GP algorithm is run on 30 splits of the data. 

\paragraph{Patient Stratification}
Each survival model is simplified into a patient stratification by generating initial patient groups and merging those with similar survival according to the log‑rank test. For survival trees, the initial groups correspond to the leaf nodes, and merging reunites leaves that actually represent the same underlying survival distribution.

PISA provides insights at three levels: (1) global insights through feature\-importance indications and feature‑set analysis, (2) model‑level insights, as each survival model is fully inspectable, and (3) patient‑stratification insights, obtained by converting each model into simple flowchart‑based stratifications.

\section{Methodology}

\subsection{Survival Tree Learning Algorithms}
Survival trees are binary trees that recursively partition the patient cohort. Survival is ultimately predicted by the survival function of the patients within each leaf node. In this work, the survival functions are built with the non-parametric Kaplan-Meier estimator~\cite{kaplan1958nonparametric}. To learn survival trees, we investigate three survival tree learning approaches: greedy survival tree (greedy ST)~\cite{leblanc1993survivaltree,wright2017ranger}, optimal survival tree (Optimal ST)~\cite{huisman2024optimal}, and evolutionary survival tree (evolutionary ST). The first two can be learned either from original covariates or from features constructed using GP-GOMEA. When GP-GOMEA constructed features are used, we shall indicate this as GFC. Using GP-GOMEA, the evolutionary ST is learned by jointly evolving the tree structure as well as the decision logic for splitting.


\subsubsection{Greedy Survival Tree (Greedy ST)} The first survival tree considered is the greedy ST~\cite{leblanc1993survivaltree} implemented in C++ by \cite{wright2017ranger}.
Iteratively, a patient set is split according to the immediate best possible split. Specifically, the split that results in the largest separation of survival subgroups according to the log-rank statistics~\cite{mantel1966evaluation} is selected. An example of a greedy ST can be seen in Figure \ref{fig:all_Trees}.

In multitree GP-GOMEA, the output of GP-Trees can be configured to construct numeric features or constrained to only construct binary features. In the former case, the learned expression of a GP-Tree can be any non-linear transformation of the input features, e.g. $BMI=weight/height^2$. Within greedy ST using GFC, a node in the survival tree then uses these numeric features and locally searches for a threshold to use in order to split the patient cohort. In the latter case, GP-GOMEA is forced to create an expression that immediately represents a split (0 or 1), e.g., $(weight/height^2)<20$. Then, in the survival tree, no thresholds need to be assigned anymore. 

\subsubsection{Optimal Survival Tree (Optimal ST)} Recently, Optimal STs were proposed~\cite{huisman2024optimal} using dynamic programming. The learning procedure, given binary input features and size constraints, can be proven to result in optimal trees. As this method is restricted to binary inputs, we constrain the features that are evolved by GP-GOMEA in this case to be binary. In contrast to the features engineered for the greedy ST, the binary input features, here, can be seen as already including both features, as well as the threshold used for the split. Finally, instead of using the same log-rank splitting rule used in greedy ST, the optimal ST selects splits that maximise the partial likelihood according to the Cox Proportional Hazards model. 

\subsubsection{Evolutionary Survival Tree (Evolutionary ST)}
Finally, we implement a new method that jointly optimises survival tree structure and decision logic using GP-GOMEA. The joint optimisation uses a multi-tree representation, in which each GP-tree is constrained to have a binary output, i.e. to capture the entire split - both feature and threshold. To build the survival tree, the GP-trees in the multi-tree representation are assigned to the survival tree structure in a breadth-first manner. That is, the first GP-tree always maps to the root node, the second GP-tree maps to the left child of the root node, the third GP-tree maps to the right child of the root node, and so on. Moreover, when traversing the survival tree, whenever a node would result in a split leading to a patient group with fewer than the hyperparameter $min\_samples\_leaf$, that subtree is pruned. An example of an evolutionary ST can be seen in Figure \ref{fig:joint tree}.

Compared to the original multi-tree GP-GOMEA~\cite{sijben2022multi}, we propose two ways to make evolving survival trees more effective: GP-tree swapping and a new initialisation strategy. The enhancements are described below, whereas their impact on evolving survival trees is discussed in the appendix \ref{appendix}. 

\paragraph{GP-Tree Swapping}
In the original multi-tree multi-objective GP-GOMEA, variation does not involve changing the ordering of GP-trees. When evolving survival trees, however, the order of GP-trees has meaning, as each GP-Tree is mapped to a fixed node in the survival tree. Therefore, adding variation to the GP-tree order is beneficial here. As such, during the variation of each individual, besides applying GP-GOMEAs variation (GOM) to every GP-tree, for each GP-tree in the multi-tree, a random GP-tree from anywhere in the multi-tree of a random donor is selected. The individual inherits this GP-tree and, if it does not deteriorate fitness, the inheritance is accepted.

\paragraph{Initialisation Strategy}
A good initialisation strategy should result in sufficient diversity (both in terms of syntax and semantics) in the initial population. Due to the limited depth of the survival trees and the pruning strategy, many individuals can lead to the same patient stratification (and thus have the same fitness). We therefore propose an initialisation strategy that resamples an individual that is similar to a previously sampled individual until it represents a different patient stratification or until a predetermined budget of tries is exhausted. That is, up to a given budget, each individual (survival tree) is forced to result in unique patient groups. To trade off the additional cost of ensuring this uniqueness and the benefit of different patient groups, different budgets are discussed in the appendix~\ref{appendix}.

\subsection{Fitness}
\label{fitness}
Both the GFC for greedy and optimal STs, as well as evolving the evolutionary ST, are done multi-objectively. Namely, we consider the survival tree performance, as well as the complexity of the features used. 

The complexity of a feature set is the sum of the size of the features that appear in the survival tree. For feature construction with greedy ST, features can be used multiple times with different thresholds and are only counted once in the complexity objective. Further, the inherent $\leq threshold$ within the greedy STs is not included in the complexity. The binary features for an optimal ST can also be reused and are only counted once. In contrast, every reuse of a feature in evolutionary STs is accounted for individually. That is because we evolve both the structure and the features multi-objectively, and thus consider the survival tree size in the complexity objective. 

The predictive performance is measured using the integrated Brier score (IBS), which describes the integral of different time-dependent Brier scores~\cite{graf1999assessment,park2021review}. Specifically, we implement the time-dependent Brier score inspired by scikit-survival~\cite{sksurv}, which measures the differences between actual survival time and survival prediction (via the survival function in a tree's leaf), and includes the inverse probability of censoring weights, a weighting scheme that accounts for censored patients. We calculate the IBS over the earliest up to the latest time in the test cohort, given that all test times are within the time range of the training patients who experienced the event.

As the performance of a survival analysis model depends largely on the split of data it is trained and tested on, we use the interquartile mean of a 25-fold stratified shuffle split. For the complexity within GFC, this means that any feature which was used in any of the 25 survival trees is considered in the complexity calculation.

\section{Showing the Need for Non-linear Feature Construction}
In this section, we demonstrate the need for non-linear feature construction when building survival trees. For this, we propose a particular synthetic survival problem, in which the true patient groups can only be efficiently identified using splits based on non-linear feature combinations.

\begin{figure}[htbp]  
    \begin{subfigure}[t]{0.4\textwidth}
        \vspace{0pt}
        \centering
        \raisebox{-0cm}{\includegraphics[width=\linewidth]{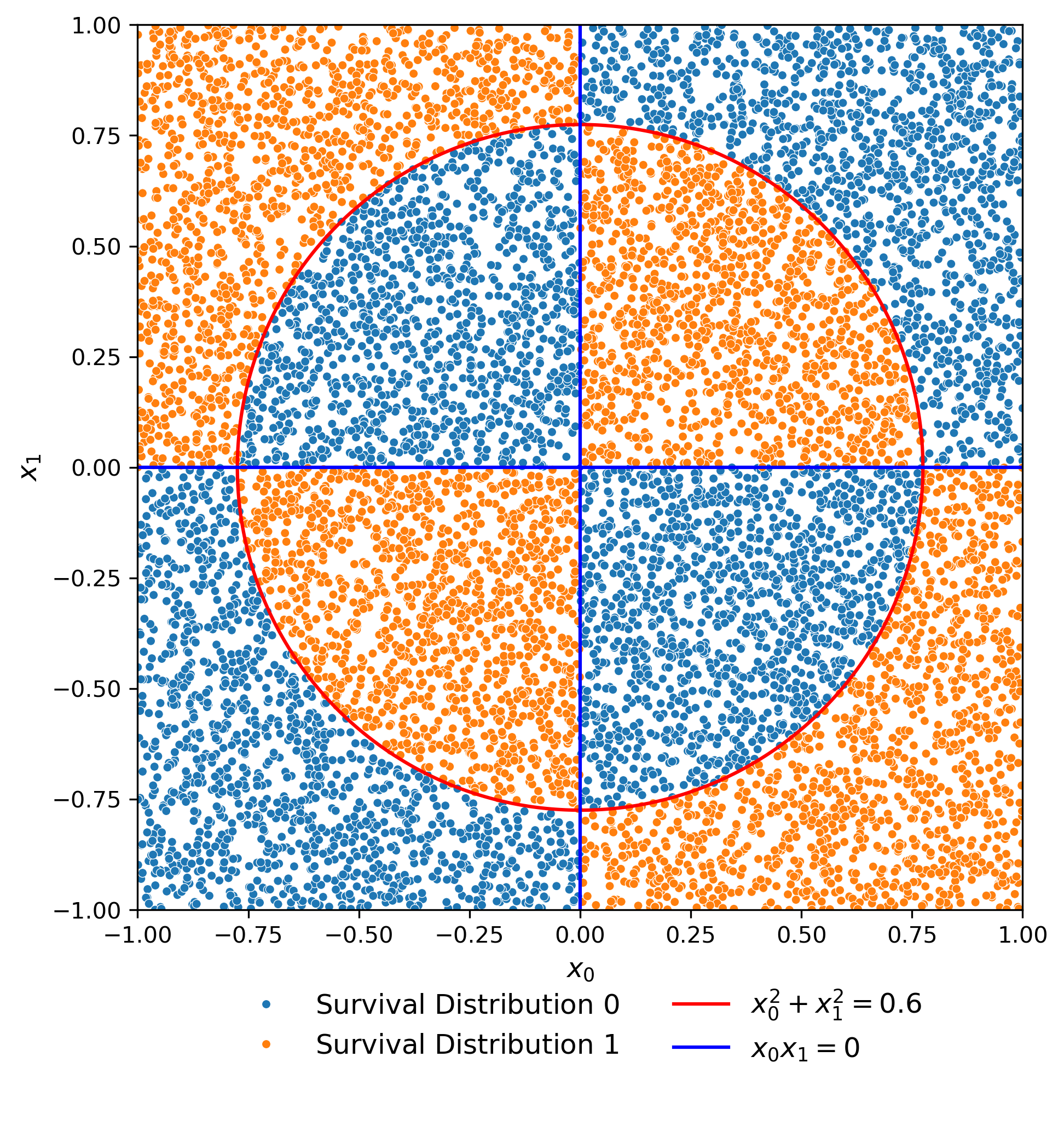}}
        \label{fig:xorgroups}
    \end{subfigure}
    \hfill
    \begin{subfigure}[t]{0.4\textwidth}
        \vspace{0pt}
        \centering
        \includegraphics[width=\textwidth]{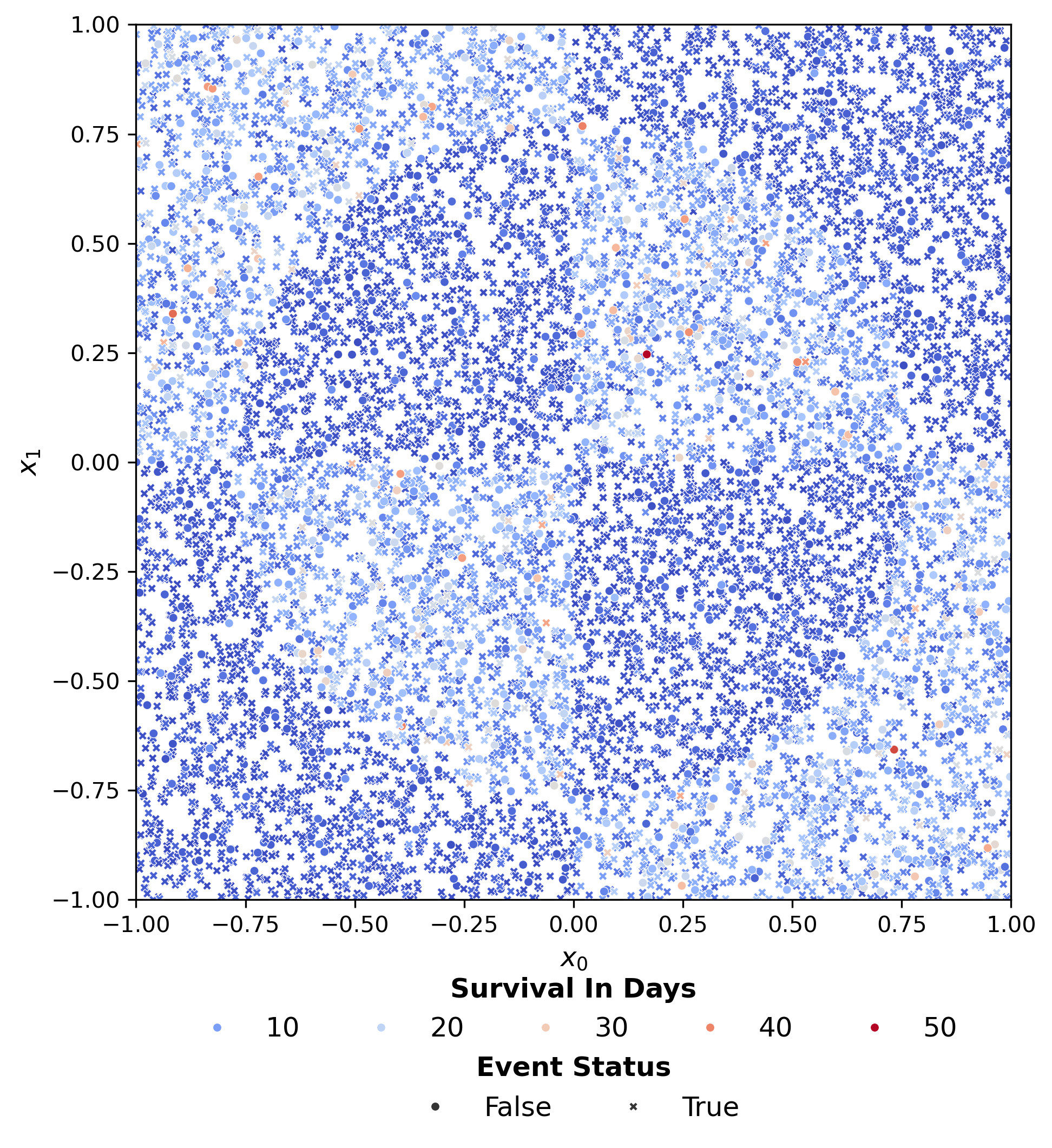}
        \label{fig:xortimes}
    \end{subfigure}
    \vspace*{-7mm}
    \caption{Survival data for the synthetic problem based on the XOR equation introduced above. Left: Data points in orange and blue and decision boundaries in red and blue. Right: survival time and event status for the datapoints.}
    \label{fig:XOR}
\end{figure}

\paragraph{Synthetic XOR Survival Data}
To generate the survival data, two survival distributions are considered, namely an exponential and a gamma distribution. Right-censoring is imitated by randomly uniformly creating censoring times, resulting in a censoring rate of 20\% and 10\% per distribution. The distributions are separated using the following formula:
$
(x_0^2+x_1^2\leq0.6)\oplus (x_0*x_1\leq0)
$, where $\oplus$ is the XOR operator.
The synthetic survival problem is dependent on two real-valued variables $x_0$ and $x_1$, which are shown along with the survival distribution assignment in Figure~\ref{fig:XOR} (left). A survival tree should thus first split on $(x_0^2+x_1^2\leq0.6)$ in the root node and subsequently split on $ (x_0*x_1\leq0)$ in each subtree, or vice versa. Each of the resulting 4 survival groups has one quarter of the patients. Crucially, the separation in the root node of an ideal tree does not result in the maximal separation of survival groups, as the resulting subgroups are made up of half of each survival distribution. That is, both splits need to be considered together. Moreover, splitting only on $x_0$ and $x_1$ requires many splits to approximate the boundaries of the distributions, as such splits are axis-parallel. 

\paragraph{Experimental Settings} For this experiment, 10-fold stratified shuffle splits are made to split the data into training and test groups of 5000, respectively. Because we are only interested in whether the survival problem can be precisely modelled with compact models, we reduce the feature construction search space to the operators needed to solve the problem, as shown in Table \ref{tab:xor_survival_results}. Further, due to the large dataset size, the 25-fold repetition within the fitness calculation is omitted. Finally, since we are concerned with whether the problem can be solved efficiently, we only report the performance of the best-performing model.

\paragraph{Findings} 
Table \ref{tab:xor_survival_results} reports the mean IBS and standard deviation over the 10 different data splits. 
As expected, the greedy ST with the original features $x_0, x_1$ cannot represent the ground truth distribution. When the correct numeric features $f_0, f_1$ are provided, it still fails to recover the ground truth because none of the required feature-threshold combinations yield the best separation of the full cohort at the root node level. Consequently, the greedy ST always selects a suboptimal split at the root node. 

Only when the two binary indicators whose interaction determines the outcome (i.e. $f_0\leq0.6$; $f_1\leq0$) are passed to the greedy ST can the ground truth be recovered. However, when additional numeric features are available ($x_0, x_1$ in this case), these may provide a larger immediate gain in the first split. In that case, the algorithm will prioritise these numerical features and fail to capture the crucial interaction between the binary indicators, resulting in a suboptimal model. Hence, only when the provided features are restricted to the correct binary indicators ($f_0\leq0.6$; $f_1\leq0$), the greedy ST can always successfully partition the cohort. 

The same conclusions can be drawn when feature construction is applied. If the operator space excludes binary operators and, thus, does not allow for finding the necessary binary indicators, then the ground truth cannot be recovered. Only when binary features can be constructed can the ground truth be found by using the greedy ST. The optimal ST with GFC and the evolutionary ST always find the ground truth, but it should be mentioned that these cannot work without binary features.


\begin{table}[t]
\centering
\renewcommand{\arraystretch}{1.2}
\setlength{\tabcolsep}{2pt}
\scalebox{0.85}{
\begin{tabularx}{1.1\textwidth}{p{15mm}|X|X|X|X}
\toprule
\textbf{Model} 
& \multicolumn{4}{c}{\textbf{Greedy ST}} \\
\cmidrule(lr){2-5}
\textbf{Features} 
& $x_0; x_1$ 
& $f_0; f_1$ 
& $x_0, x_1$;\linebreak $f_0\leq0.6$; $f_1\leq0$ 
& $f_0\leq0.6$; $f_1\leq0$  \\
\midrule
\textbf{IBS} 
& 0.106 $\pm$ 0.003
& 0.096 $\pm$ 0.003
&0.083 $\pm$ 0.016
& \textbf{0.074 $\pm$ 0.003} \\
\bottomrule
\end{tabularx}}

\scalebox{0.85}{
\begin{tabularx}{1.1\textwidth}{p{15mm}|X|X|X|X}
\toprule
\textbf{Model} 
& \multicolumn{2}{c|}{\textbf{Greedy ST (GFC)}} 
& \textbf{Optimal ST (GFC)} 
& \textbf{Evolved ST}\\
\cmidrule(lr){2-5}
\textbf{Features} 
& Engineered\linebreak ($+, *, ^2$)
& Engineered ($+, *, ^2, \leq$) 
& Engineered ($+, *, ^2, \leq$) 
& Engineered ($+, *, ^2, \leq$) \\
\midrule
\textbf{IBS} 
& 0.092 $\pm$ 0.003
& \textbf{0.074 $\pm$ 0.003} 
& \textbf{0.074 $\pm$ 0.003}
& \textbf{0.074 $\pm$ 0.003} \\
\bottomrule
\end{tabularx}}
\caption{Mean Integrated Brier Score (IBS) with standard deviation across 10 data splits. $f_0\leq0.6$ and $f_1\leq0$ are binary covariates where $f_0=x_0^2+x_1^2$ and $f_1=x_0*x_1$. Constructed features are built using the specified operators. The ground-truth survival tree has an IBS of $0.074\pm0.003$.}
\label{tab:xor_survival_results}
\end{table}

\section{Experimental Set-up}
\subsection{Clinical Benchmark Datasets}

We use two real-world clinical datasets that both concern survival in patients with breast cancer. We derived the data from \cite{katzman2018deepsurv,knottenbelt2025coxkan}, so that we can directly compare to the state-of-the-art performances of DeepSurv~\cite{katzman2018deepsurv} and CoxKAN~\cite{knottenbelt2025coxkan}.

\paragraph{GBSG}
The GBSG dataset, pre-processed by \cite{katzman2018deepsurv} and \cite{altman2000we}, facilitates the overall survival prediction of patients with node-positive breast cancer. 1546 patients (of which 37\% are censored) are taken from the Rotterdam Tumour Bank~\cite{foekens2000urokinaseROTTERDAMGBSG} and are used as the internal validation set, whereas the external validation set consists of 696 patients (with a censoring rate of 56\%) from Germany~\cite{schumacher1994randomizedGBSGGERMAN}. The covariates include hormone therapy status, the tumour size (categorised into 3 groups), the patient's menopausal status, as well as their age, the number of involved lymph nodes, and the progesterone (PGR) and oestrogen (ER) receptor levels measured in fentomoles per litre.

\paragraph{METABRIC} 
The Molecular Taxonomy of Breast Cancer International Consortium (METABRIC) dataset, preprocessed by \cite{katzman2018deepsurv},  allows for the survival prediction of patients with breast cancer~\cite{curtis2012genomic}. In METABRIC 20\% of the same cohort is held out as the external validation set. This results in 1523 patients in the internal dataset and 381 patients in the external validation set, both with 57\% censored patients. METABRIC includes information on 4 gene indicators, as well as information on whether a patient underwent hormone-, radio- or chemotherapy, a patient's age and whether the patient is ER-positive, i.e. whether the tumour has an oestrogen receptor~\cite{katzman2018deepsurv}.

\subsection{Hyperparameters}
The hyperparameters chosen are taken from the hyperparameters used in the multiple feature multi-objective feature engineering step in PISA. That is, for each of the clinical datasets, we run GP-GOMEA using the different tree induction methods 30 times on different splits of the internal cohort. We use the same population size of 1024. A run is either terminated at a maximum of 50 generations or whenever the hypervolume has not improved in the last 5 generations. The hypervolume~\cite{zitzler2002multiobjective} is a commonly used measure to describe the quality of an approximation set in a multi-objective space, given a reference point (for which we use the maximum possible feature set size and the worst possible IBS score, 1). Following PISA's motivation to promote the chances of interpretability in the proposed feature sets, the operator set is selected to include relatively simple operators. This is in contrast to other state-of-the-art methods, e.g. CoxKAN, which allows complex operators, for instance, trigonometric functions. Finally, GP-GOMEA uses a fixed GP-tree template, which we limit to a depth of 3 in this work. All survival trees have a minimum of 2 \% samples per leaf. The evolutionary ST uses tree swapping and a unique group assignment budget of 1000 at initialisation. 

Besides the deep learning-based baselines DeepSurv~\cite{katzman2018deepsurv} and CoxKan~\cite{knottenbelt2025coxkan}, as well as the Random Survival Forest (RSF)~\cite{ishwaran2008random}, which all either do not offer the interpretability of individual survival trees (DeepSurv, RSF) or have the downside of adhering to the proportional hazard assumption (DeepSurv, CoxKan), we compare the greedy ST with and without GFC, the optimal ST with and without GFC, and the evolutionary ST. Following the insights from the synthetic survival problem, we include two configurations for the greedy STs with GFC, one using only numeric operators and one including binary operators. As the optimal ST induction using the original features requires binary features, we binarise the features as described in their work~\cite{huisman2024optimal}. As a final baseline, we consider the non-parametric Kaplan-Meier estimator, which builds one survival function across the entire cohort. That is, every patient gets the same survival prediction depending on time only. A summary of the hyperparameters of the configurations used in this work can be seen in Table \ref{tab:hyperparameters}.

\begin{table}
    \centering
    \hspace*{-3mm}
    \scalebox{0.82}{
    \begin{tabular}{c|c|c|c|c}
        \textbf{ST Learning Method} & \textbf{ST Depth}  & \makecell{\textbf{\# Engineered}\\\textbf{Features}} & \makecell{\textbf{Binary GP-Tree}\\ \textbf{output constraint} } &\textbf{Operator Set}\\
        \midrule
        Greedy ST (GFC) &  2 &  $\leq$3&False & $+,-,*,^2,/$\\
        Greedy ST (GFC) &  3 &  $\leq$3&False &$+,-,*,^2,/$\\
        Greedy ST (GFC) &  2 &  $\leq$3&False &$+,-,*,^2,/,\leq,NOT,AND,OR$\\
        Greedy ST (GFC) &  3 &  $\leq$3&False &$+,-,*,^2,/,\leq,NOT,AND,OR$\\
        Optimal ST (GFC) &  2 &  $\leq$3&True&$+,-,*,^2,/,\leq,NOT,AND,OR$\\
        Optimal ST (GFC) &  3 &  $\leq$7&True&$+,-,*,^2,/,\leq,NOT,AND,OR$\\
        Evolutionary ST &  2 &  $\leq$3&True&$+,-,*,^2,/,\leq,NOT,AND,OR$\\
        Evolutionary ST &  3 &  $\leq$7&True&$+,-,*,^2,/,\leq,NOT,AND,OR$\\
        \bottomrule
    \end{tabular}}
    \caption{The different configurations and their hyperparameters used to construct features and learn survival trees using GP-GOMEA.}
    \label{tab:hyperparameters}
\end{table}

\subsection{Performance Metrics}
Performance is evaluated on the (held-out) external validation test set. Specifically, we bootstrap the test set to get performance estimates, from which we empirically (via 1000 resamples) derive the mean and its 95\% confidence interval. As a metric, we use IBS, as described in Section \ref{fitness}, as well as Harrell's right-censored Concordance Index (C-index)~\cite{harrell1996multivariable,sksurv}. Whereas the IBS provides estimates on the calibration and discrimination of the survival functions, the C-index, which allows for straightforward comparison across methods as it does not rely on a time range like the IBS does, measures the discrimination of the predictions. That is, the C-index counts the number of correctly ranked patients in terms of their survival prediction and their actual survival. 

\section{Results \& Discussion on Clinical Datasets}
\begin{figure}[htbp]  
    \begin{subfigure}[t]{0.32\textwidth}
        \centering
        \includegraphics[width=\linewidth]{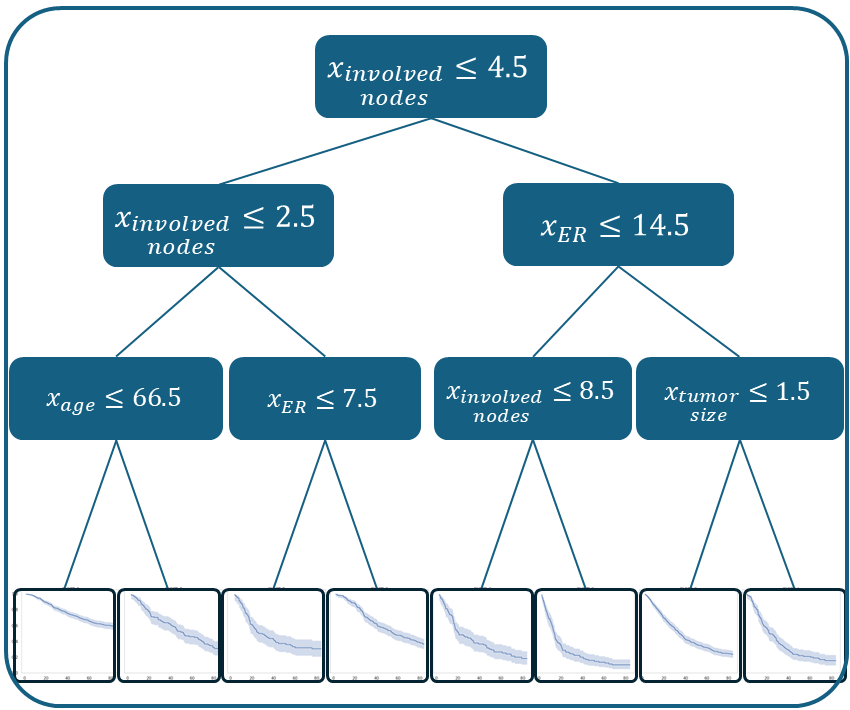}
        \caption{}
        \label{fig:baseline_st}
    \end{subfigure}
    \begin{subfigure}[t]{0.32\textwidth}
        \centering
        \includegraphics[width=\textwidth]{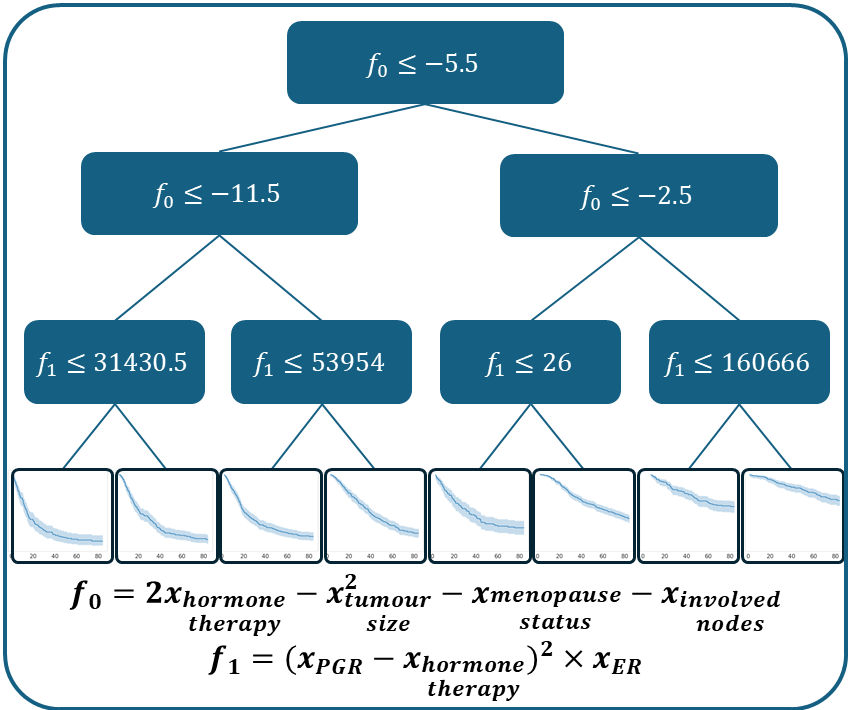}
        \caption{}
        \label{fig:engineered features}
    \end{subfigure}
        \begin{subfigure}[t]{0.32\textwidth}
        \centering
        \includegraphics[width=\textwidth]{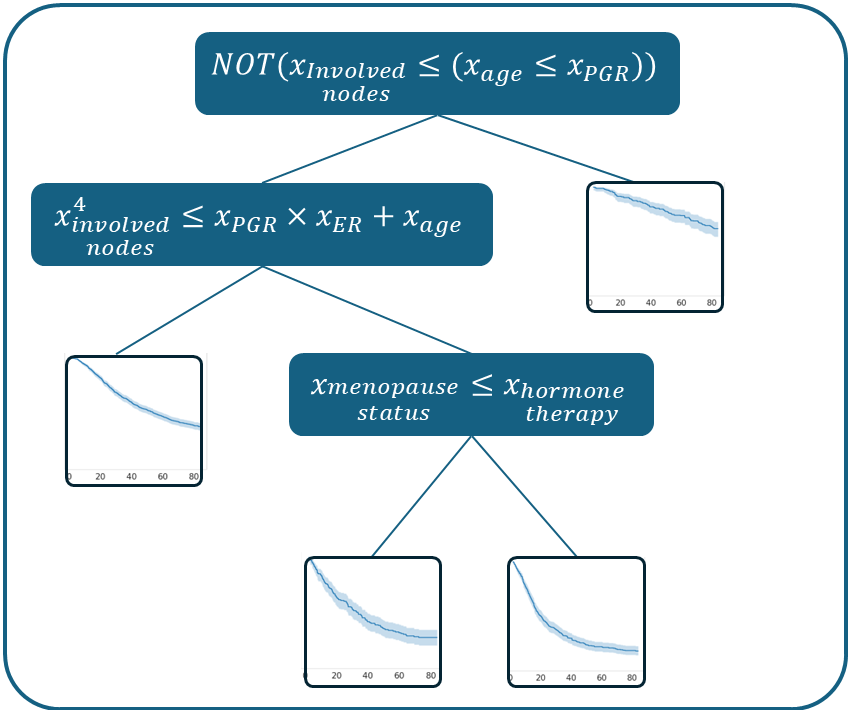}
        \caption{}
        \label{fig:joint tree}
    \end{subfigure}
    \caption{\textbf{Top performing survival trees of depth 3 for GBSG.}  Subfigure \ref{fig:baseline_st} shows a greedy ST on the original input features, achieving a mean IBS of 0.124, whereas Subfigure \ref{fig:engineered features} shows the GFC greedy ST of depth 3 that achieves the lowest mean IBS of 0.115. Subfigure \ref{fig:joint tree} shows the evolutionary ST of depth 3, in which both features and tree structure are evolved jointly, which achieves the lowest mean IBS of 0.116. Note that here the engineered binary features include the cut-offs, whereas in the greedy STs, threshold values are determined during ST building.}
    \label{fig:all_Trees}
\end{figure}

\begin{figure}[htbp]
    \centering
    \includegraphics[width=1\linewidth]{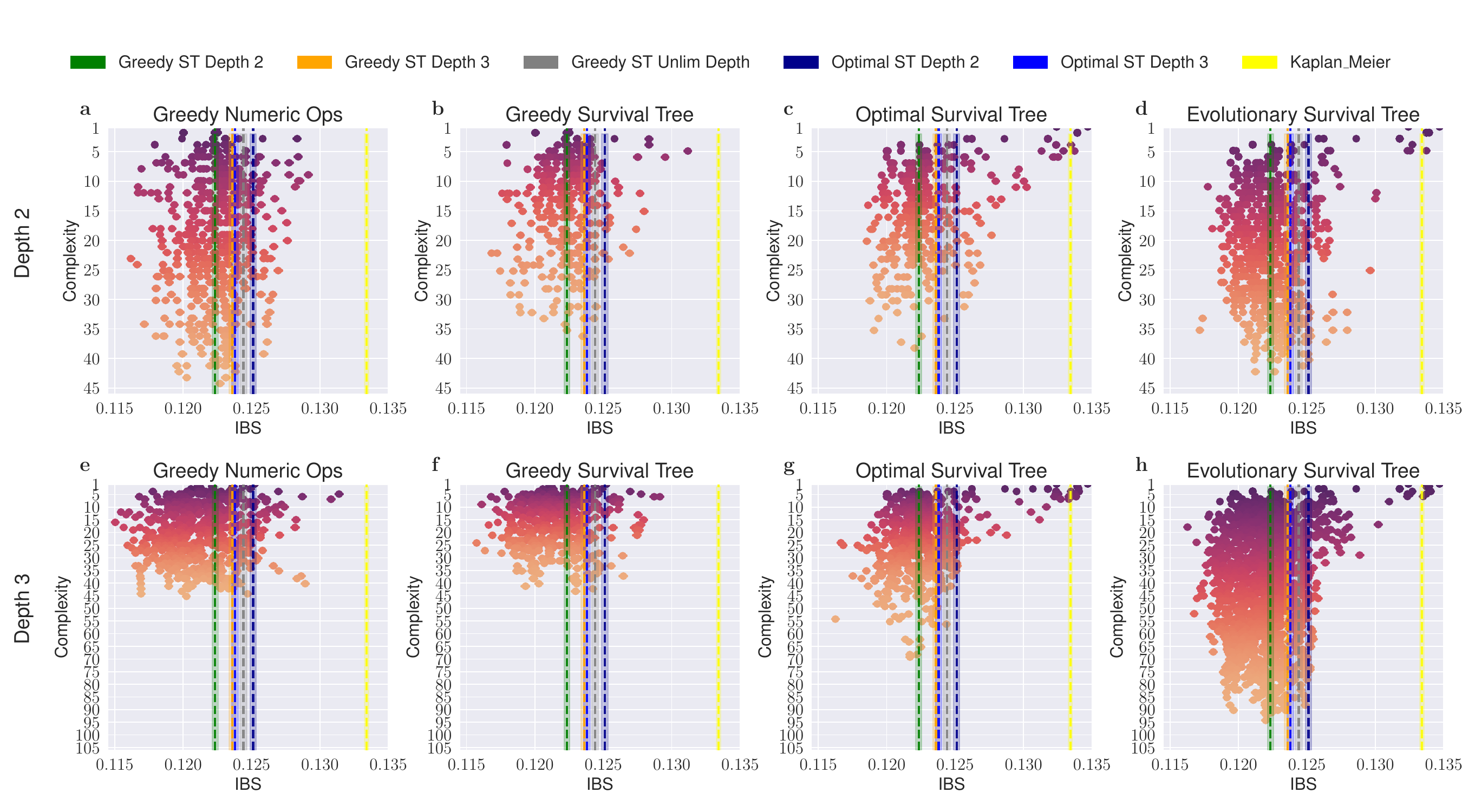}
    \caption{\textbf{IBS performances on the GBSG use case} The bootstrapped mean IBS and its 95\% confidence interval over 1000 bootstraps on the external cohort. All dots are survival tree models derived via multiple-feature multi-objective construction, whereas the lines represent baselines made on the original input features.}
    \label{fig:gbsg_full_performances}
\end{figure}

\begin{figure}[htbp]
    \centering
    \includegraphics[width=1\linewidth]{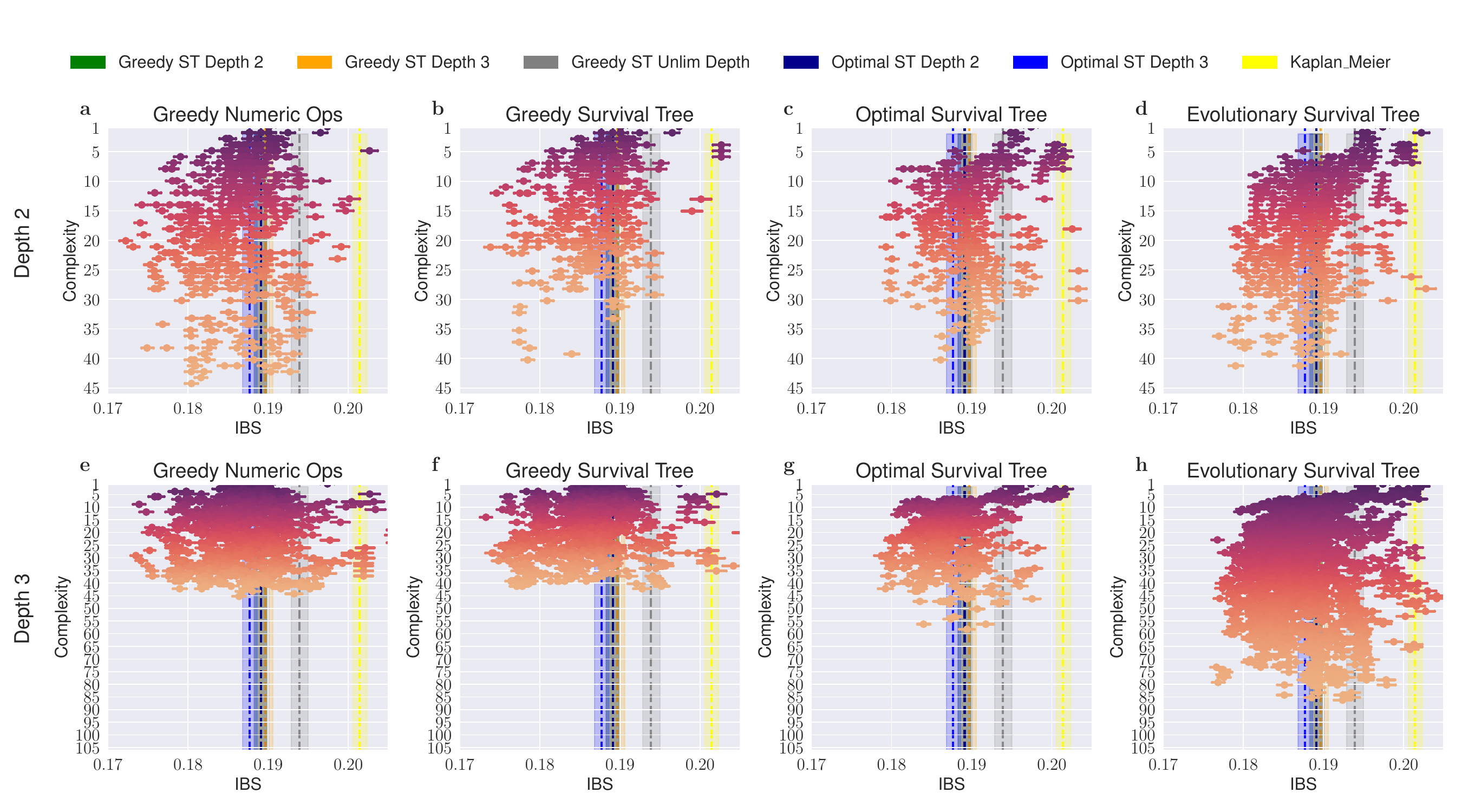}
    \caption{\textbf{IBS performances on the METABRIC use case.} Bootstrapped mean IBS and 95\% confidence interval over 1000 bootstraps on the external cohort. Dots are ST models derived via multiple-feature multi-objective construction, whereas lines represent baselines made using original input features.}
    \label{fig:metabric_full_performances}
\end{figure}

Figures \ref{fig:gbsg_full_performances} and \ref{fig:metabric_full_performances} show that the survival trees using engineered features outperform the respective baselines, especially when using survival trees of depth 3. The baseline greedy ST, as well as the top-performing models for the greedy ST with GFC and numeric operators and the evolutionary ST are shown in Figure \ref{fig:all_Trees}, demonstrating their interpretability.

Surprisingly, the GFC greedy ST with no binary operators performs on par with the other approaches. This shows that in use cases in which multiple jointly optimal splits and general binary operations are not necessary to model the survival prediction well (in contrast to the XOR case tested earlier), using a reduced numeric operator set for GFC can still lead to good survival trees even when using a greedy learning algorithm. 

Moreover, when using GFC for greedy STs, the evolutionary search is not restricted to finding binary features, but may evolve numeric features. The greedy ST learning algorithm then locally searches for the best splits. In contrast, the other survival tree induction approaches need to fully evolve a binary feature (composited covariate and threshold) within a GP-tree, essentially solving a harder problem. On top of this, as all features in GP-GOMEA use a GP-tree depth of 3, the optimal and evolutionary ST approaches must capture more information in the GP-trees (composited covariate and threshold), essentially giving the GP-trees in the GFC greedy ST more expressive power.

Greedy STs, with or without GFC, may repeatedly split on the same feature at different thresholds, as seen in Figure \ref{fig:baseline_st} and \ref{fig:engineered features}. Reusing features is particularly interesting as it can improve interpretability, since clinicians are required to understand fewer distinct features. While the reuse of features can also be done with the evolutionary STs approach, the current complexity calculation would count each reuse of a feature, whereas for the GFC approaches, any reused feature is only counted once. While this means that within the evolutionary ST the survival tree size itself is optimised multi-objectively, it does not discount the reuse of features, meaning that comparing complexity metrics across the GFC survival trees and evolutionary ST can be misleading. 

The evolutionary ST also performs well and has several advantages. Specifically, using the evolutionary STs makes fitness evaluation computationally efficient as the trees are directly induced from their genotype and can be readily evaluated, whereas combinations with greedy or optimal ST building approaches require running an exogenous procedure first. To run the first generation of the GFC with greedy ST on only numeric operators, the GFC with greedy ST with all operators, the GFC on the optimal ST, and the evolutionary ST (all with 3 GP-trees and a survival tree depth of 3) on one core, takes 19 hours, 10 hours, 2.1 hours, and 45 minutes, respectively. Note that in practice, the configurations are run in parallel and experience significant speedups.
In addition to the increased computational efficiency, by evolving the entire split within the nodes, the evolutionary ST can induce locally sub-optimal splits easily that still constitute an overall optimal survival tree (as seen within the synthetic survival problem).

In Table \ref{tab:placeholder}, we compare the results of the survival trees to well-known traditional and state-of-the-art methods in terms of their C-index. From this, we see that using feature construction combined with the various tree induction techniques, survival trees can perform on par with state-of-the-art approaches, while not needing the Cox proportional hazards assumption and using interpretable features to split the patient cohort into a maximum of 8 groups.


\begin{table}[tbh]
\renewcommand{\arraystretch}{1.2}
\setlength{\tabcolsep}{5pt}
\centering
\scalebox{0.9}{
\begin{tabularx}{\textwidth}{X|X|X|X|X|X|X|X}
         Dataset & \textbf{Greedy ST} \tiny Depth 3 & \textbf{Optimal ST} \tiny Depth3 & \textbf{RSF} & \textbf{Cox PH}  & \textbf{Deep-Surv*} & \textbf{Cox-KAN*} & \textbf{Cox-KAN*} \tiny Symbolic   \\
         \midrule
         \textbf{GBSG} 
         & 0.643 \tiny(0.64, 0.647)  
         & 0.639 \tiny(0.636, 0.643) 
         & 0.669 \tiny(0.666, 0.673) 
         & 0.653 \tiny(0.650, 0.657) 
         &0.668 \tiny(0.616, 0.620)  
         & 0.678 \tiny(0.676, 0.682) 
         &0.683 \tiny(0.678, 0.684)  \\
         \textbf{META-BRIC }
         & 0.606 \tiny{(0.602, 0.610)} 
         & 0.62 \tiny{(0.616, 0.624)} 
         & 0.630 \tiny{(0.625, 0.635)} 
         & 0.631 \tiny{(0.626, 0.636)} 
         & 0.643 \tiny{(0.64, 0.648)} 
         & 0.647 \tiny(0.644, 0.652)  
         & 0.650 \tiny(0.644, 0.651) \\
         \bottomrule
    \end{tabularx}}
\scalebox{0.9}{
\begin{tabularx}{\textwidth}{X|X|X|X|X|X|X|X|X}
         Dataset &  \multicolumn{2}{|X|}{\textbf{Greedy ST}\tiny GFC\_numeric\_ops} & \multicolumn{2}{|X|}{\textbf{Greedy ST}\tiny GFC\_all\_ops}& \multicolumn{2}{{|X|}} {\textbf{Optimal ST}\tiny GFC} &  \multicolumn{2}{|X}{\textbf{Evolutionary ST}} \\
         ST Depth 
         & 2
         & 3
         & 2
         & 3
         & 2
         & 3
         & 2
         & 3 \\
         \midrule
        \textbf{GBSG }
         & 0.667 \tiny(0.665, 0.67)
         & 0.684 \tiny(0.681, 0.687)
         & 0.663 \tiny(0.66, 0.666)
         & 0.674 \tiny(0.671, 0.677)
         & 0.664 \tiny(0.662, 0.667)
         & 0.672 \tiny(0.67, 0.675)
         & 0.671 \tiny(0.668, 0.674)
         & 0.676 \tiny(0.67, 0.679) \\
         \textbf{META-BRIC }
         & 0.635 \tiny(0.631, 0.64)
         & 0.657 \tiny(0.653, 0.661)
         & 0.631 \tiny(0.626, 0.635)
         & 0.663 \tiny(0.659, 0.667)
         & 0.636 \tiny(0.632, 0.64)
         & 0.649 \tiny(0.645, 0.654)
         & 0.639 \tiny(0.635, 0.643)
         & 0.661 \tiny(0.656, 0.665)\\

         \bottomrule
    \end{tabularx}}
    \caption{C-index comparison between survival trees (ST) with engineered features and well-known survival analysis methods, as well as state-of-the-art approaches, both using the original input features. The performances are given by bootstrapping the same external test set 100 times and then are summarised via the sample mean and its 95\% confidence interval given 1000 bootstrapped samples. For survival trees using the engineered features, the best-performing model is chosen. * indicates that the results were taken from their original work (\cite{katzman2018deepsurv} and \cite{knottenbelt2025coxkan}); however, the external set is the same across all methods. The different configurations of survival tree approaches using feature construction can be seen in 
    Table \ref{tab:hyperparameters}. The C-index of a single Kaplan-Meier survival curve is 0.5, as every patient gets the same survival prediction and is ranked equally to each other.}
    \label{tab:placeholder}
\end{table}

\section{Conclusion}
The advantage of learning survival trees for survival analysis is that there is no need for the Cox proportional hazards assumption. Moreover, if the survival tree is sufficiently shallow, and the decision logic sufficiently easy to understand, the entire model can be considered transparent, verifiable, and traceable, which are important aspects for real-world use.
This work demonstrates that effective feature construction is central to constructing powerful, yet shallow survival trees. By adopting a multi-objective feature construction approach, we enable the generation of multiple shallow survival trees that offer alternatives for users to inspect with different trade-offs between complexity and accuracy. Further, on two clinical datasets, the use of feature construction for a greedy ST learning approach, as well as a fully evolutionary ST learning approach, resulted in survival trees that achieve predictive performance on par with state-of-the-art methods that do use the Cox proportional hazards assumption.

While already competitive, in the current implementation, fully evolutionary STs have limitations. In particular, binary features need to be evolved that constitute decisions in the survival tree. Likely, it would be beneficial to reuse numerical features and optimise thresholds using evolutionary methods to have more flexibility without requiring higher complexity both in the representation and the search space, as features then no longer need to be independently discovered in separate locations in the survival tree. This would merge the advantages of feature reuse with non-greedy survival tree building and the efficient evaluation-only fitness computation that a genetic programming procedure offers. A modular representation of GP-GOMEA, as seen in \cite{harrison2025thinking}, could provide such reusable features, which we consider a promising direction for future work.

Further, for uniformity, we have used the same population size, maximum GP-tree depth, and maximum generations across the approaches. However, as GP-GOMEA solves a more complex problem in the evolutionary ST case, future research should investigate whether increasing the resources for the evolutionary ST approach would aid in increasing performance further.

Finally, this work demonstrates that the key to powerful shallow survival trees is multi-objective feature construction, while multi-objectively evolving both the tree structure and split logic likely holds the most promise given its speed and flexible representation. 
\begin{credits}
\subsubsection{\ackname} 
We thank Kenzo Boudier for their preliminary exploratory analysis on this idea. This research is part of the "Uitlegbare Kunstmatige Intelligentie" project funded by the Stichting Gieskes-Strijbis Fonds. Furthermore, we thank NWO for the Small Compute grant on the Dutch National Supercomputer Snellius.
\end{credits}

\bibliographystyle{splncs04}
\bibliography{PPSN_bib}
\newpage
\appendix
\begin{figure*}[htbp]
\begin{subfigure}[t]{0.5\textwidth}
    \centering
    \includegraphics[width=\textwidth]{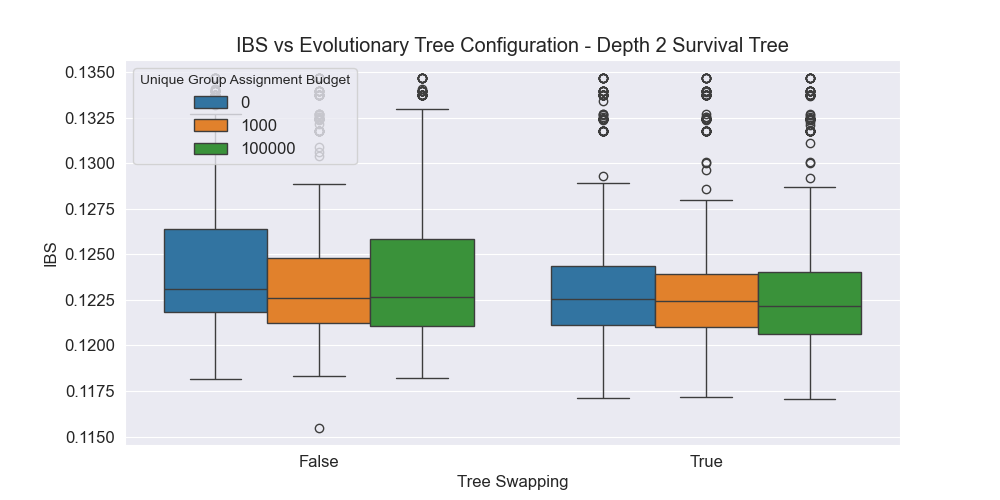}
    \caption{}
    \label{IBS_3trees}
\end{subfigure}
\begin{subfigure}[t]{0.5\textwidth}
    \centering
    \includegraphics[width=\textwidth]{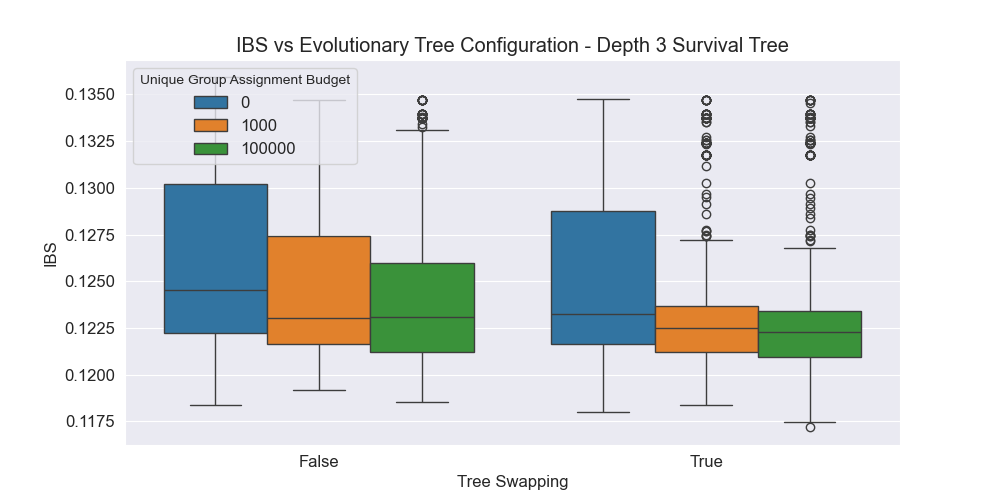}
    \caption{}
    \label{IBS_7trees}
\end{subfigure}
\begin{subfigure}[t]{0.5\textwidth}
    \centering
    \includegraphics[width=\textwidth]{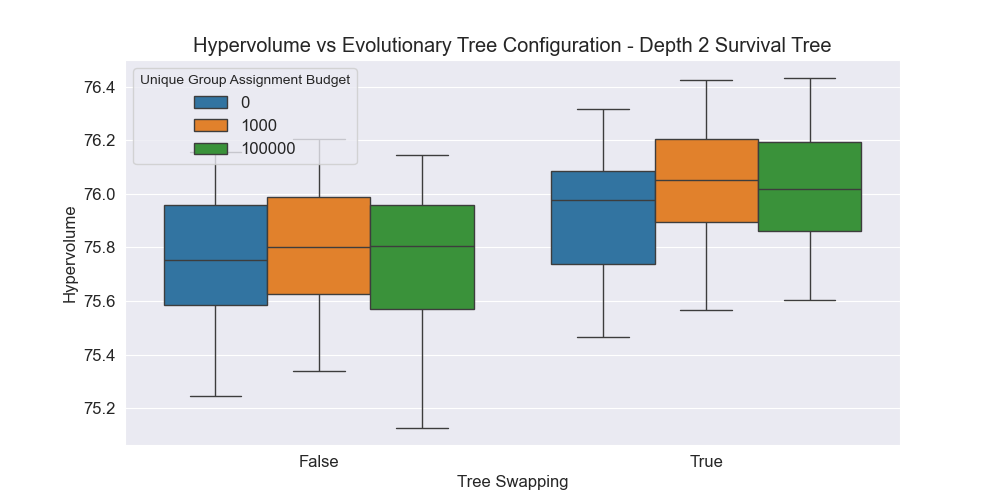}
    \caption{}
    \label{Hypervolume_3trees}
\end{subfigure}
\begin{subfigure}[t]{0.5\textwidth}
    \centering
    \includegraphics[width=\textwidth]{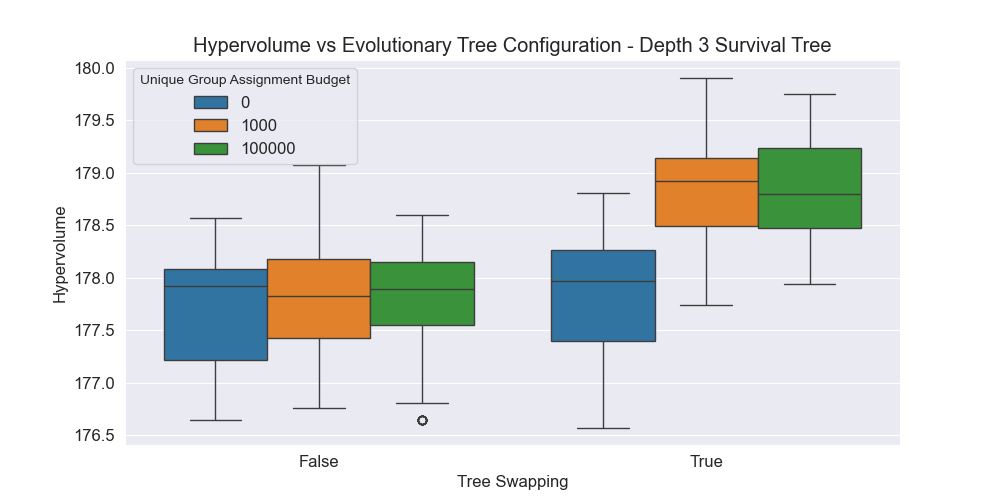}
    \caption{}
    \label{Hypervolume_7trees}
\end{subfigure}
\caption{\textbf{IBS performances and achieved hypervolume across different configurations for the evolutionary STs of depth 2 and 3.}}
\label{fig:ablation_study_metrics}
\end{figure*}
\begin{figure}[htbp]
\begin{subfigure}[t]{\textwidth}
    \centering
    \includegraphics[width=0.8\textwidth]{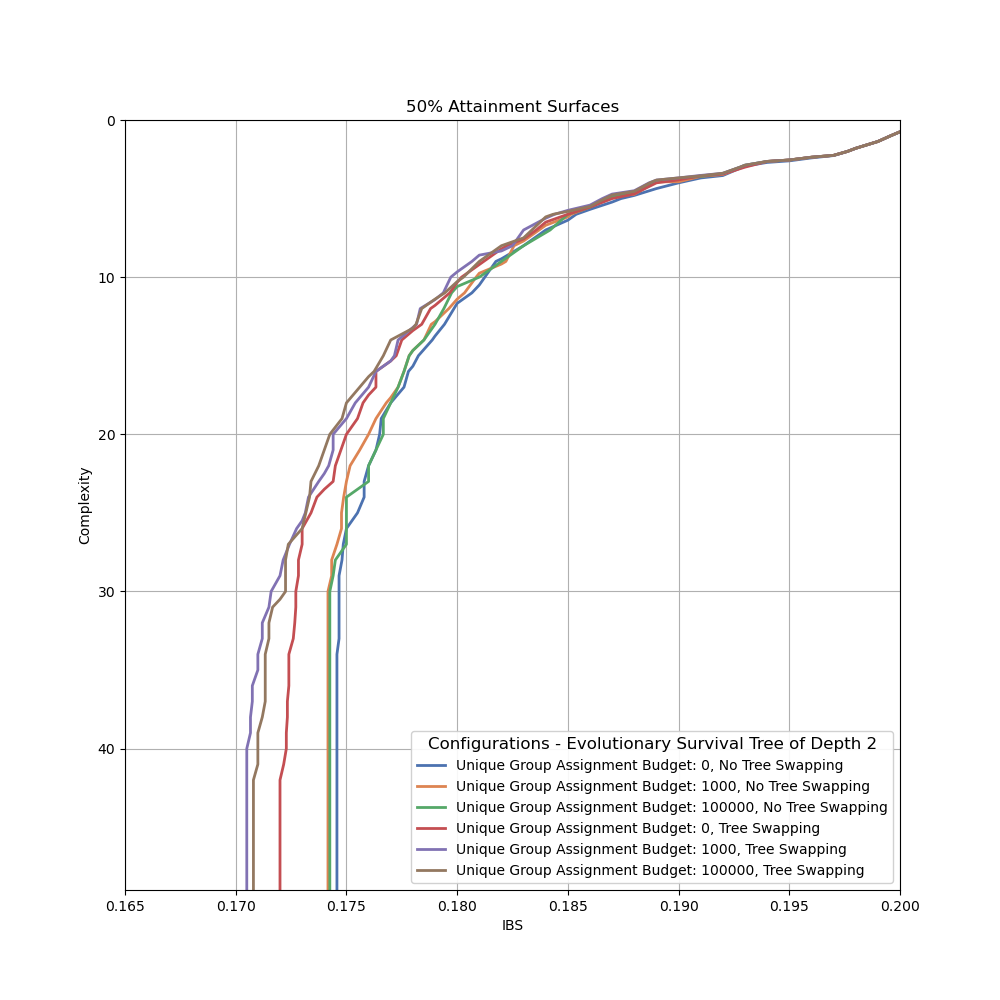}
    \caption{}
    \label{fig:Attainment_3trees}
\end{subfigure}
\begin{subfigure}[t]{\textwidth}
    \centering
    \includegraphics[width=0.8\textwidth]{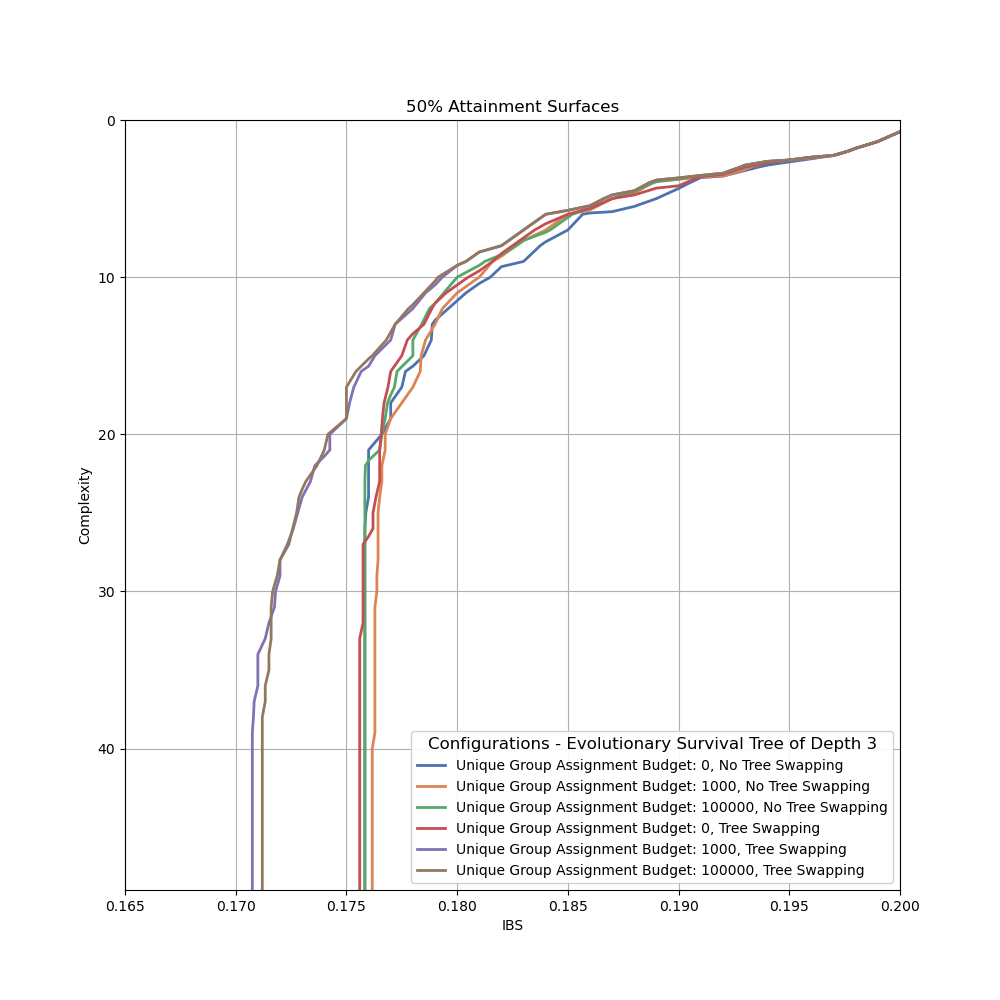}
    \caption{}
    \label{fig:Attainment_7trees}
\end{subfigure}
\caption{\textbf{50\% Attainment surfaces for the different evolutionary ST configuration} Figure \ref{fig:Attainment_3trees} shows the 50\% attainment surface on different configurations of the evolutionary ST depth of 2, whereas Figure \ref{fig:Attainment_7trees} shows the 50\% attainment surface on different configurations of the evolutionary ST depth of 3.}
\label{fig:attainmentsurfaces}
\end{figure}
\section{Ablation study of the improvements for the evolutionary ST}
\label{appendix}
In this section, the enhancements designed to enhance the evolutionary ST are briefly investigated. For this, we use the GBSG use case and run 6 different configurations: with and without tree swapping, combined with three budgets for ensuring unique group assignment at initialisation: 0, 1000, 100000. A unique group assignment budget of 0 no individual is resampled to be different to another in terms of their patient stratification, whereas a budget of 1000 accounts for roughly more than half of the patients to have unique group assignments. Finally, a budget of 100000 ensures that all individuals have unique group assignments.

The evolutionary ST is run 30 times on different data splits on two different survival tree depths: a survival tree depth of 2 uses 3 GP-trees of depth 3, whereas the survival tree depth of 3 uses 7 GP-trees with a depth of 2.

The performances of the different models using the different configurations can be seen in Figure \ref{IBS_3trees} and \ref{Hypervolume_3trees} for survival trees of depth 2, i.e. 3 GP trees, and in Figure \ref{IBS_7trees} and \ref{Hypervolume_7trees} for survival trees of depth 3, i.e. 7 GP trees. Figure \ref{IBS_3trees} and \ref{IBS_7trees} show the spread of IBS performances for each model found in the 30 repetitions, whereas Figure \ref{Hypervolume_3trees} and \ref{Hypervolume_7trees} show the hypervolume for each of the 30 runs. Observing these results indicates that adding tree swapping aids in performance, specifically when combined with some budget to initialise the population with to capture groups. 

To visualise where hypervolume improvements occur, Figures~\ref{fig:Attainment_3trees} and~\ref{fig:Attainment_7trees} show the 50\% attainment plots~\cite{fonseca1996performance} for the six configurations at depths~2 and~3. The 50\% attainment surface marks the boundary that at least half of the optimisation runs managed to reach or dominate, making it a useful indicator of where consistent performance gains appear. The largest improvements occur at higher model complexities, but the advantage of tree swapping is already visible at lower complexities. When seven trees are used, the improvement becomes even more pronounced, as swapping has a greater effect when more trees are available. 
\end{document}